\documentclass[runningheads]{llncs}

\usepackage{eccv2024template/eccv}

\usepackage{eccv2024template/eccvabbrv}

\usepackage[dvipsnames]{xcolor}

\newcommand{\xhdr}[1]{\vspace{5pt}\noindent\textbf{#1}}

\newcommand{\cmark}{\ding{51}}%

\definecolor{graycolor}{gray}{.9}

\newcommand{\ourmodel}{GMRW}

\usepackage{amsmath,amsfonts,bm}

\def\eqref#1{equation~\ref{#1}}

\def\1{\bm{1}}

\DeclareMathAlphabet{\mathsfit}{\encodingdefault}{\sfdefault}{m}{sl}
\SetMathAlphabet{\mathsfit}{bold}{\encodingdefault}{\sfdefault}{bx}{n}

\def\gL{{\mathcal{L}}}

\def\sR{{\mathbb{R}}}

\newcommand{\E}{\mathbb{E}}

\usepackage{graphicx}
\usepackage{booktabs}
\usepackage{amsmath}
\usepackage{colortbl}
\usepackage{multirow}
\usepackage{arydshln}
\usepackage{comment}
\usepackage{enumitem}
\usepackage{tabularx}
\usepackage{float}

\newcommand{\qb}[0]{{\mathbf q}}
\newcommand{\pb}[0]{{\mathbf p}}
\newcommand{\corr}[0]{F}

\newcolumntype{L}[1]{>{\raggedright\let\newline\\\arraybackslash\hspace{0pt}}m{#1}}
\newcolumntype{C}[1]{>{\centering\let\newline\\\arraybackslash\hspace{0pt}}m{#1}}
\newcolumntype{R}[1]{>{\raggedleft\let\newline\\\arraybackslash\hspace{0pt}}m{#1}}

\usepackage[accsupp]{axessibility}  %

\definecolor{eccvblue}{rgb}{0.12,0.49,0.85} %

\usepackage[pagebackref,breaklinks,colorlinks,citecolor=eccvblue]{hyperref}

\usepackage{amsmath,amsfonts,amssymb} %
\usepackage{pifont}%

\usepackage{orcidlink}

\begin{document}

\title{Self-Supervised Any-Point Tracking by Contrastive Random Walks}

\titlerunning{GMRW}

\author{Ayush Shrivastava \and
Andrew Owens}

\authorrunning{A.~Shrivastava et al.}

\institute{University of Michigan\\
\email{\{ayshrv,ahowens\}@umich.edu} \\
\vspace{1em}\url{https://ayshrv.com/gmrw}
}

\maketitle

\begin{abstract}

We present a simple, self-supervised approach to the Tracking Any Point (TAP) problem. We train a global matching transformer to find cycle consistent tracks through video via  contrastive random walks, using the transformer's attention-based global matching to define the transition matrices for a random walk on a space-time graph. The ability to perform ``all pairs'' comparisons between points allows the model to obtain high spatial precision and to obtain a strong contrastive learning signal, while avoiding many of the complexities of recent approaches (such as coarse-to-fine matching). To do this, we propose a number of design decisions that allow global matching architectures to be trained through self-supervision using cycle consistency. For example, we identify that transformer-based methods are sensitive to shortcut solutions, and propose a data augmentation scheme to address them. Our method achieves strong performance on the TapVid benchmarks, outperforming previous self-supervised tracking methods, such as DIFT, and is competitive with several supervised methods.

\end{abstract}
\section{Introduction}

The problem of finding space-time correspondences underlies a number of computer vision tasks. An emerging line of work on the Tracking Any Point (TAP) problem~\cite{doersch2022tap} has addressed the specific challenges of tracking over long time horizons: estimating all future and past positions of any given physical point in a video. This problem addresses the shortcomings of traditional formulations of long-range tracking, such as chained optical flow and sparse tracking, enabling applications in animation~\cite{doersch2022tap} and robotics~\cite{vecerik2023robotap}. Yet the difficulty of acquiring labeled training data has restricted the capabilities of these models. Existing models are thus limited to training on small, synthetic datasets.

\begin{figure*}[t]
    \centering
    \includegraphics[width=\textwidth]{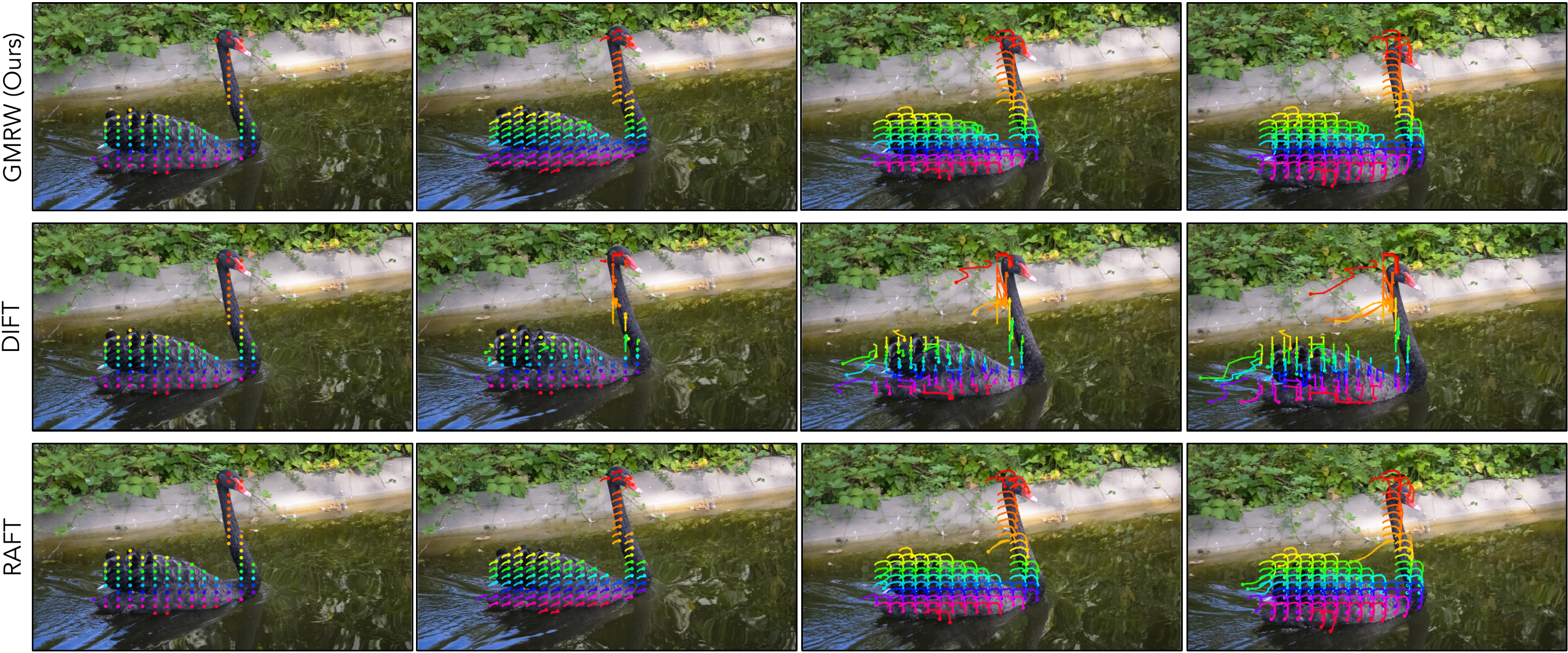}
    \caption{{\bf Global Matching Random Walks.} We present a self-supervised method for tracking all physical points over the course of a video, i.e., the Tracking Any Point problem~\cite{doersch2022tap}. Our model uses a global matching transformer~\cite{xu2022gmflow} to track points cycle consistently over time, using the contrastive random walk~\cite{jabri2020space}. Our approach outperforms self-supervised tracking methods, such as self-supervised DIFT~\cite{tang2023emergent} and supervised optical flow methods, like RAFT~\cite{teed2020raft}, on the TAP-Vid benchmark~\cite{doersch2022tap}.}
    \label{fig:teaser_figure}
\end{figure*}

This is in contrast to many other areas of computer vision~\cite{he2020momentum,he2022masked,doersch2015unsupervised}, where 
self-supervised methods have arisen as a powerful way to learn from unlabeled data.  While a number of such methods have been proposed for space-time correspondence~\cite{vondrick2018tracking,wang2019learning,jabri2020space,bian2022learning,xu2021rethinking}, these methods are not well suited to the challenge of tracking physical points over long time horizons, the core challenge of Tracking Any Point (Fig.~\ref{fig:teaser_figure}). Self-supervised optical flow methods~\cite{jonschkowski2020matters,bian2022learning,yu2016back}, for example, obtain dense short-range motion fields but struggle to track over long time horizons, while methods that excel at semantic tracking, such as recent methods that repurpose text-to-image diffusion features~\cite{tang2023emergent}, match together points that belong to the same object category, not necessarily those that correspond the same physical points.

In this paper, we propose a simple and effective self-supervised approach to the Tracking Any Point problem. %
We adapt the global matching transformer architecture~\cite{xu2022gmflow} to learn through cycle consistency~\cite{zhou2016learning,wang2019learning,jabri2020space}: i.e., tracking forward in time, then backward, should take us back to where we started. In lieu of labeled data, we supervise the model via the contrastive random walk~\cite{jabri2020space}, using the self-attention from global matching to define the transition matrix for a random walk that moves between points in adjacent frames. This ``all pairs'' matching mechanism allows us to define transition matrices that consider large numbers of points at once, thereby increasing spatial precision and enabling us to obtain a richer learning signal by considering a large number of paths through the space-time graph on which the random walk is performed. Additionally, we identify that global matching architectures are susceptible to shortcut solutions (e.g., due to their use of positional encodings), and that previously proposed methods for addressing these shortcuts are insufficient~\cite{tang2021breaking}. We therefore propose a type of data augmentation that removes these shortcuts. 

Our approach obtains strong performance on the TAP-Vid~\cite{doersch2022tap} benchmark,  significantly outperforming previous self-supervised tracking methods on TAPVid-DAVIS and Kubric. Through experiments, we show:
\begin{itemize}[label=$\bullet$,itemsep=0pt,topsep=0pt,leftmargin=5mm]
\item Self-supervised models can obtain strong performance on the Tracking Any Point task, e.g., obtaining competitive performance to TAP-Net~\cite{doersch2022tap} on many metrics.
\item The contrastive random walk can successfully be extended to long-range point tracking.
\item Global matching transformers can be trained through cycle consistency.
\item Data augmentation can remove shortcut solutions to cycle consistent training.
\end{itemize}

\section{Related Work}

\xhdr{Space-time representation learning.} A variety of recent methods have been proposed for learning to track pixels through video via self-supervision. Vondrick et al.~\cite{vondrick2018tracking} learned a representation in which pixels were photoconsistent, providing the model with only grayscale images during training and using the held-out colors to assess the quality of the match. While this approach is effective, it implicitly assumes that objects are photo-consistent over time, an assumption that is frequently violated~\cite{jonschkowski2020matters}. Another line of work proposes to use cycle consistency. Wang et al.~\cite{wang2019learning} used a model based on spatial transformers to track pixels forward in time, then backwards, learning a representation that minimized the distance from their point of origin. Other work has combined these two approaches together~\cite{li2019joint} or use two-stage matching~\cite{lai2020mast}. Jabri et al.~\cite{jabri2020space} formulated video cycle consistency as a random walk on a graph containing space-time patches, providing dense supervision to the model. We extend this approach and use a transformer architecture that allows for global matching and finer correspondences. Bian et al.~\cite{bian2022learning} showed that the contrastive random walk can obtain pixel-accurate matches through multi-scale matching, and proposed several extensions that unify it with self-supervised optical flow methods~\cite{jonschkowski2020matters}. In addition to focusing on obtaining spatially precise matches, our goal is to predict tracks in long videos. However, we note that multi-scale matching could be combined with our method to obtain more accurate results. Tang et al.~\cite{tang2021breaking} proposed an extension that allowed for fully convolutional training, avoiding shortcut solutions. They proposed a data augmentation technique to use different image crops for the forward and backward cycles of the random walk. However, this does not avoid shortcut solutions for transformer-based methods. Recent work has learned features that change slowly over time~\cite{xu2021rethinking,gordon2020watching} or by adding temporal alignment~\cite{hadji2021representation},  and other work has proposed standardized tracking benchmarks~\cite{mckee2022transfer}. Another recent work~\cite{gupta2024siamese} extends the Masked Autoencoders~\cite{he2022masked} to videos. They mask large fractions of patches in the video and learn visual representations by reconstructing the missing patches.

\xhdr{Optical flow.} A parallel line of work has focused on creating unsupervised models for optical flow~\cite{jason2016back, ren2017unsupervised, wang2018occlusion, liu2019ddflow, jonschkowski2020matters, stone2021smurf}. Typically these models combine simple photometric losses (e.g., using hand-crafted features) with heavy augmentation and physical constraints, such as smoothness. RAFT~\cite{teed2020raft} proposed a recurrent architecture that updates the flow field through iterative all-pairs matching and regression. Different from prior flow approaches, GMFlow~\cite{xu2022gmflow} formulated optical flow prediction as a global matching problem that identifies correspondences by directly comparing feature similarities. Extending GMFlow, Xu et al.~\cite{xu2023unifying} proposed a unified model for flow, rectified stereo matching and unrectified stereo depth estimation from posed images. Rather than predicting two-frame velocity estimates which most optical flow methods do, we learn probabilistic matches between frames. To do this, we adopt the transformer-based global matching network architecture of Xu et al.~\cite{xu2022gmflow}, which performs non-parametric matching. However, instead of supervising the model to generate flow estimates, we train it to perform a contrastive random walk.  

\xhdr{Cycle consistency.} Zhou et al.~\cite{zhou2016learning} proposed to use cycle consistency as a supervisory signal across different instances of the same category to train correspondence models. Dwibedi et al~\cite{dwibedi2019temporal} used temporal cycle consistency between multiple varying videos to learn representations useful for fine-grained temporal understanding in videos. Other methods have used cycle-consistency to detect occlusions~\cite{sun2021autoflow, lei2009optical, hur2017mirrorflow, baker2011database}, such as within unsupervised flow models~\cite{janai2018unsupervised, jonschkowski2020matters, wang2018occlusion,zou2018df}. Other work uses cycle consistency for semi-supervised learning~\cite{haeusser2017learning}. ~\cite{zhao2021modelling} represents the video as two sub-graphs, one which connects inter-frame nodes (similar to Jabri et al.~\cite{jabri2020space}) and another which connects intra-frame nodes located in local neighborhood and use cycle-consistency to perform random walk on this graph. 

\xhdr{Tracking.} Recently, many works have focused on supervised methods for long-term pixel tracking along with the introduction of new benchmarks. TAP-Vid~\cite{doersch2022tap} proposed a new test-bed for tracking any point in a video. They released four benchmarks based on real and synthetic videos for evaluating tracking methods. They also provided a tracking model called TAP-Net that compares the features of a query point with all frames to predict tracks independent of time. Concurrently, Persistent Independent Particles (PIPs) ~\cite{harley2022particle} introduced a tracking method that searches over a local neighborhood and smooths the estimates over time by iterative refinement. They also release a dataset called FlyingThings++ based on FlyingThings~\cite{mayer2016large}. Neoral et al~\cite{neoral2024mft} proposed a supervised tracker that exploits optical flows from consecutive frames and pairs of frames at logarithmically scaled intervals. TAPIR~\cite{doersch2023tapir} uses a two-stage approach: matching stage that provides an initial location of the query point in every frame and a refinement stage that updates the tracks based on local neighborhoods, essentially combining techniques from TAP-Net and PIPs models.  CoTracker~\cite{karaev2023cotracker} introduced a transformer architecture that jointly tracks multiple points throughout an entire video. OmniMotion~\cite{wang2023tracking} introduces a test-time optimization method by building globally consistent motion representations. All these methods use ground-truth tracks or trajectories computed from pretrained optical flow methods for training, whereas we use self-supervision to train our model provided by cycle-consistency as the supervisory signal.

\section{Method: Global Matching Random Walk}

\begin{figure*}[t]
    \centering
    \includegraphics[width=\textwidth]{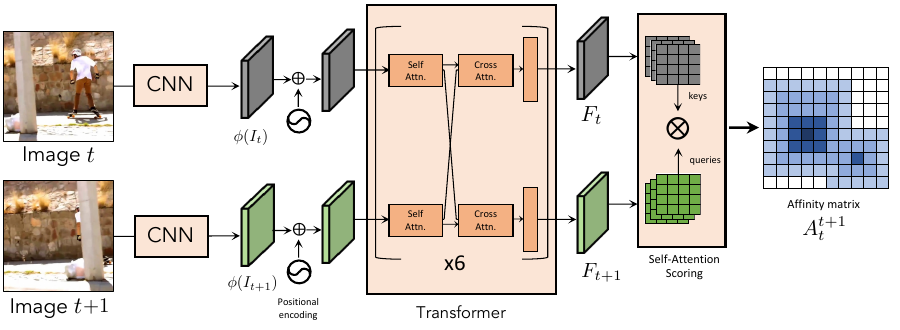}
    \caption{{\bf Model Architecture.} Our model takes a pair of images $I_t$ and $I_{t+1}$ as input over which it computes correspondences. We extract visual features from a CNN, add positional encodings, and pass them as tokens to our global matching transformer. The transformer consisting of 6 stacked layers of self-attention, cross-attention and feed-forward networks, processes these features and produces correlated features $F_t$ and $F_{t+1}$. We compute self-attention over $F_t$ and $F_{t+1}$ and use the attention as the transition matrix for performing contrastive random walks. To compute tracks during evaluation, we can take an expectation over the affinity matrix to get coordinates $(x, y)$.}
    \label{fig:model_architecture}
\end{figure*}
We propose a self-supervised method for tracking points in video. We use the contrastive random walk~\cite{jabri2020space} to learn cycle-consistent track, using an architecture based on self-attention from a global matching transformer. Our use of a transformer-based architecture, which requires addressing additional shortcut solutions that were not present in previous work on cycle consistent tracking, which we address through an augmentation scheme. We train on unlabeled videos and perform point tracking in a strided fashion to get pixel-level features for matching.

\xhdr{Problem setup.} 
In the Tracking Any Point (TAP) problem, we are given a video $V$ with frames $\{I_t\}_{t=1}^T$ where $I_t \in \sR^{H \times W \times 3}$ and query points $\{\qb_k\}_{k=1}^N$  where each $\qb_k=(t_k, x_k, y_k)$ represents a timestep in the video and its position in the frame $I_{t_k}$. The goal is to produce point trajectories, $\pb_t^k=(x_t^k, y_t^k)$ for all $t\in T$ for each query point $\qb_k$. In addition to positions, the visibility  $v_t^k \in \{0, 1\}$ indicate whether the point $\pb_t^k$ is occluded in frame $I_t$. Note that the query point can be provided for any timestep in the video, not necessarily the first frame, and that at that timestep the query point is assumed to be visible. 

Following recent supervised approaches~\cite{doersch2022tap}, we create a model that operates on a pair of (not necessarily temporally adjacent) video frames. At test time, this model can be used to track points over long time horizons, either by directly matching pairs of frames or by chaining.

\subsection{Global matching architecture} \label{sec:architecture} %
In order to obtain spatially precise correspondences, we require an architecture (Fig.~\ref{fig:model_architecture}) that can perform efficient, global matching. We adapt the recent GMFlow architecture of Xu et al.~\cite{xu2022gmflow,xu2022unifying} for self-supervised tracking. This architecture has previously been applied to optical flow and stereo-matching in supervised settings. 

A key advantage of this design is that it finds a match by ``all pairs'' matching through self-attention. This is in contrast to alternative architectures, which rely on regressing motion~\cite{teed2020raft,harley2022particle,sun2018pwc,doersch2023tapir} and thus provide only a single point estimate. We use global matching to define a transition matrix for the contrastive random walk, which allows us to model the motion of a very large number of points at once. This has several advantages to previous approaches to the contrastive random walk~\cite{jabri2020space,bian2022learning}. First, the motion can be captured at a much finer-grained level. For example, in our experiments we divide an $256 \times 256$ resolution image into a {$64\times64$ grid }, whereas Jabri et al.~\cite{jabri2020space} was limited to a coarse $7 \times 7$ grid and Bian et al.~\cite{bian2022learning} required coarse-to-fine matching. Second, it allows us to obtain more supervision per iteration through contrastive learning, since the model can simultaneously explore a larger number of paths through the space-time graph on which we will perform the random walk.

\xhdr{Image features.} 
We extract $d$-dimensional features $\phi (I_t)\in \sR^{\frac{H}{c}\times\frac{W}{c}\times d}$ for each image $I_t$ from a convolutional neural network where $c=4$. We add 2D positional encoding to these visual features.  While the visual encoder used in Xu et al.~\cite{xu2022gmflow} computes features at two scales, we only perform single-scale matching for simplicity.

\xhdr{Global correlation matching.}  
For each pair of consecutive video frames, $I_t$ and $I_{t+1}$, we process the image features through six layers of stacked self-, cross-attention and feed-forward networks to get correlation features, $\corr_{t}$, $\corr_{t+1}$. The keys and values for cross-attention layers come from the same feature, but queries come from the other feature in the pair. We also use Swin-style shifted local windows to improve computation efficiency~\cite{xu2022gmflow,liu2021swin}. 

\xhdr{Computing a random walk transition matrix.} 
Once we have correlation features $\corr_{t}$, $\corr_{t+1}$, we compute the transition matrix as $A_{t}^{t+1} = \text{softmax}(\corr_{t} {\corr^{\top}_{t+1}} / \tau)$. This matrix represents the probability of a patch in frame $t$ matching to frame $t+1$, and is used to define the transition probability for a contrastive random walk. In contrast to previous contrastive random walk models, which (following work in contrastive learning~\cite{wu2018unsupervised,he2020momentum,chen2020simple}) $L_2$-normalize the embedding features and apply a small temperature parameter~\cite{jabri2020space}, we follow the standard practice in transformers and use unnormalized features with a large normalization constant $\tau = \sqrt{d}$. We provide full architectural details in the supplementary material.

\xhdr{Sampling stride.} Our model uses probabilities from the affinity matrix to find correspondences. The resolution of the affinity matrix is determined by the spatial resolution of the correlation features $F_t$, $F_{t+1}$. To compute pixel-level features, we use different feature strides to sample features from the images following previous works~\cite{neoral2024mft,karaev2023cotracker}. We implement this by upsampling the original image by the stride. We vary our sampling strides among  $s=\{1, 2, 4\}$ and mention the stride used during training and evaluation in our experiments.

\xhdr{Estimating the motion and visibility.} 
Given a transition matrix $A_{s,t}$ between frames $s$ and $t$, we can compute the expected change in position, following~\cite{bian2022learning,xu2022gmflow}. In other words, we take a weighted sum of the positions of the points we match in frame $t$:
\begin{equation}
\label{eq:motion}
    \mathbf{f}_{s,t} = \mathbb{E}_{A_{s,t}} [A_{s,t} D - D],
\end{equation}
where $\mathbf{f}_{s,t} \in \mathbb{R}^{n\times2}$ is the matrix of predicted optical flows, $D \in \mathcal{R}^{n\times2}$ is the (constant) matrix containing of pixel coordinates for each point, and $A_{s,t}D$ is the expected position from frame $s$ to $t$. 

To estimate the visibility $v_i$ for a given point, we perform a cycle consistency test: we check whether the predicted motion fields $\mathbf{f}_{s,t}$ and $\mathbf{t,s}$ produce a motion that is within a threshold $\tau_{cyc}$ (we use $\tau_{cyc} = 3$ pixels) of the original point. Those that exceed this threshold are marked as being invisible.

\subsection{Learning the model}

\xhdr{Cycle-consistency by contrastive random walks.}
We use the cycle consistency objective introduced contrastive random walks in Jabri et al.~\cite{jabri2020space} to supervise our model. We treat the video as a space-time graph and train a model to perform random walks. We use our model to compute the transition matrices for the walker, which walks from patches in frame $t$ to $t+1$, then back to $t$. This sequence is known as a {\em palindrome} sequence. We supervise the model by maximizing the probability that the walker returns to its initial location, which amounts to minimizing the loss: %
\begin{equation}
\gL_{\mathtt{crw}} = \gL_{\mathtt{CE}}( {A}_{t}^{t+1}{A}_{t+1}^{t}, I),
\label{eq:crw_loss}
\end{equation}
where ${A}_{t}^{t+1}{A}_{t+1}^{t}$ is the (chained) transition matrix for the random walk, and $\gL_{\mathtt{CE}}$ is cross-entropy loss, and $I$ is the identity matrix. In other words, the model penalizes random walks whose transition matrices differ from the identity, and which thus are not cycle-consistent.

\begin{figure*}[t]
    \centering
    \includegraphics[width=\textwidth]{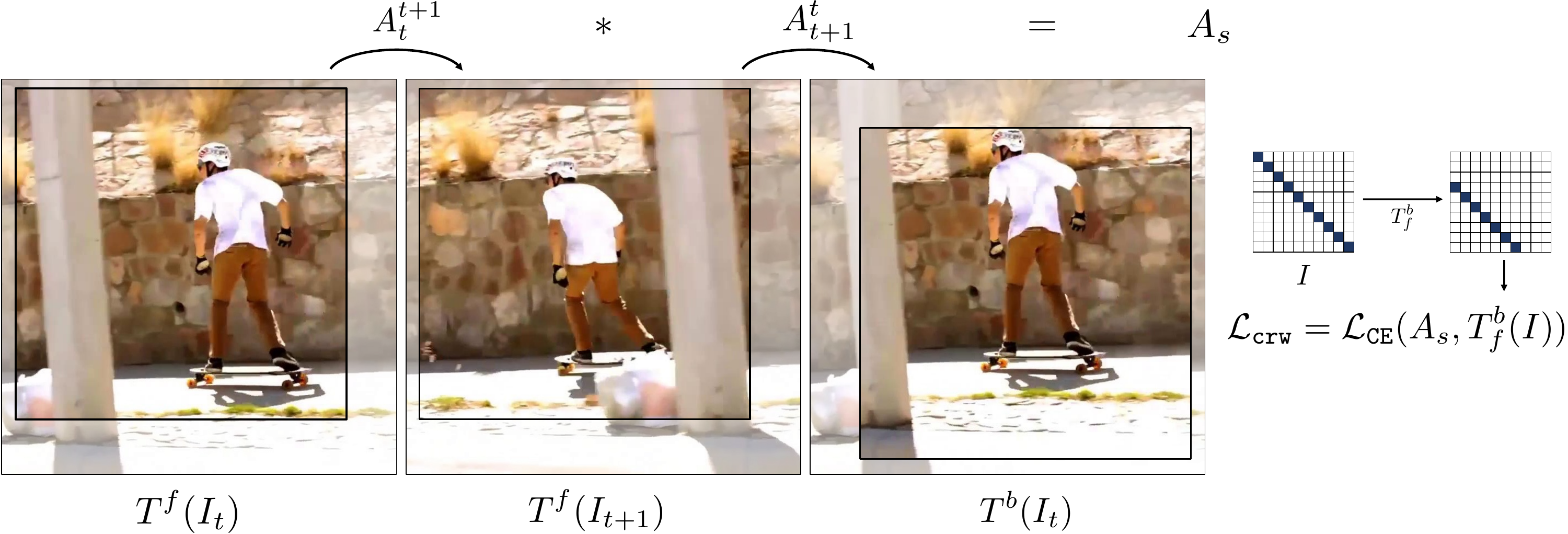}
    \caption{{\bf Label Warping.} We propose label warping as a remedy to avoid shortcut solutions that arise when we use transformer-based models for contrastive random walks. Instead of warping the last feature to match the first feature, we propose to warp the label used for cycle consistency. For an image pair $I_{t}, I_{t+1}$, we apply different affine transformations $T^f$, $T^b$ to the forward and backward cycle. We then compute $A_{t}^{t+1}$, $A_{t+1}^{t}$ and chain them together to get the affinity matrix for cycle consistency. We then supervise it with the warped identity matrix $T_{f}^{b}(I)$ where $T_f^b$ represents the transformation to go from $T^f$ to $T^b$.}
    \label{fig:labelwarp}
\end{figure*}

\xhdr{Shortcut solutions for global matching models.}
Many cycle consistency learning schemes are susceptible to a trivial solution~\cite{jabri2020space,bian2022learning}: ignore the visual content of the patch, and match solely based on its position. Since the transformer model has a positional encoding, and our architecture has a large number of global self-attention layers, this solution is easy to find. Tang et al.~\cite{tang2021breaking} showed that this shortcut could be removed in simple fully convolutional CNN models by randomly cropping and resizing the video frames, using different (but consistent) augmentations in the forward and backward directions of the walk. When computing the loss (Eq.~\ref{eq:crw_loss}), they undo the crop by warping the last feature in the affinity matrix with the inverse transformation, and supervise the affinity matrix to be the identity $I$. 
We found this method still led to shortcut solutions. In our case, the CNN features are processed by a transformer, which is capable of undoing the warp and trivially matching based on the positional information. 
To address this issue, we perform {\em label} warping (Fig.~\ref{fig:labelwarp}). Instead of warping the network features, we warp the labels to match the transformation between the forward and backward cycle. Let $T^f$ and $T^b$ be different resize-crop transformations applied to the forward and backward cycles of the contrastive random walk. Specifically, for the image sequence $(I_{1}$, $I_{2}$, $I_{1})$, we construct the frame sequence $[T^f(I_1), T^f(I_2), T^b(I_1)]$ and compute the loss: %
\begin{equation}
 \gL_{\mathtt{crw}} = \gL_{\mathtt{CE}}( A_s, T_f^b(I)),
\end{equation}
where $A_s$ is the chained transition matrices for the full random walk, and $T_f^b(I)$ denotes that we apply the same spatial transformation to the label set.

\xhdr{Smoothness loss.} We also evaluate model variations that impose spatial smoothness. To ensure the movement of random walkers is smooth, we also consider variations of the model that follow Bian et al.~\cite{bian2022learning} and add an edge-aware smoothness loss~\cite{jonschkowski2020matters}: 
\begin{equation}
    \gL_{\mathtt{smooth}} = \E_p \sum_{d \in \{x, y\}} \text{exp} (-\lambda_c I_d(p)) |\frac{\partial^2 \textbf{f}_{s,t}(p)}{\partial d^2}|   
\end{equation}
where the $p$ is a pixel, $I_d (p) = \frac{1}{3} \sum_{c}\left| \frac{\partial I_c}{\partial d}\right|$ is the spatial derivative averaged over all color channels $I_c$ in direction $d$, and where we use the second derivatives of the estimated flow field. The parameter $
\lambda_c$  controls the influence of pixels with similar colors. When both losses are used together, the model minimizes:
\begin{equation}
    \gL_{\mathtt{total}} = \gL_{\mathtt{crw}} + \lambda_s \gL_{\mathtt{smooth}},
\end{equation}
where $\lambda_s$ is a constant. %

\section{Experiments}

Our model estimates space-time correspondences between a pair of video frames, from which we can compute expected coordinates for each track. We evaluate our method on the four Tap-Vid benchmarks and use the standard evaluation metrics~\cite{doersch2022tap}.

\subsection{Training}
Following Doersch et al.~\cite{doersch2022tap}, we use the Kubric dataset~\cite{greff2022kubric} as our main dataset for training. The TapVid-Kubric dataset consists of synthetic images of a few objects randomly moving around in the video. It contains a training set of 38,325 videos of $256 \times 256$ resolution and 799 validation videos. During training, we randomly sample 2 frames separated by a few timesteps from the video and create a palindrome sequence ($I_1\rightarrow I_2\rightarrow I_1\textquotesingle$). We then apply different random resized crops to forward ($I_1, I_2$) and backward images ($I_1\textquotesingle$) and train for cycle consistency and smoothness loss. Unlike other models that address TAP, we do not use any labels from the training data, as our method is entirely self-supervised. To evaluate the ability of our model to generalize to ``in the wild'' internet video, we also train a version of the model using the Kinetics 400~\cite{kay2017kinetics} dataset using a similar 2-frame sampling strategy.

\subsection{Evaluation}
To validate our approach and to test the generalization of our trained models, we run evaluations on the TapVid benchmarks~\cite{doersch2022tap}, namely on Kubric, DAVIS, Kinetics, and RGB-Stacking.  TapVid-DAVIS is a real dataset of 30 videos from DAVIS 2017 validation~\cite{pont20172017} with videos from 34-104 frames. Similarly, TapVid-Kinetics is a real dataset of 1000+ videos from Kinetics~\cite{kay2017kinetics} with 250 frames each. TapVid-RGB-Stacking is a dataset of 50 synthetic videos with 250 frames each from robotic hand object manipulation task. We mainly use TapVid-DAVIS for our testbed as it is based on real videos. 

\xhdr{Metrics.} In TapVid benchmarks, the query points are sampled using a special strategy (when using \textit{strided} query method)~\cite{doersch2022tap} and then are tracked forward and backward in time. Performance is evaluated on: (1) the positional accuracy $\mathtt{< \delta^{x}_{avg}}$ metric for frames in which the point is visible, the fraction of points that are within a threshold over the ground truth, averaged over several thresholds. (2) \textit{Occlusion Accuracy} which is a classification accuracy for predicting where a point is visible or occluded. (3) \textit{Average Jaccard} (AJ), the fraction of true positives (points within a threshold of visible ground truth points), divided by true positives and false positives, averaged over multiple thresholds~\cite{doersch2022tap}.  

\section{Results}

We compare our method to previous supervised and self-supervised approaches, and evaluate a number of different design decisions.

\begin{table*}[ht]
\renewcommand{\arraystretch}{1.3} %
\centering
{\resizebox{1.0\linewidth}{!}{
\small
\begin{tabular}{p{3mm}lccccccccccccc}
\toprule
\multirow{2}{*}{\textbf{}} & 
\multirow{2}{*}{\textbf{Method}} & 
\multirow{2}{*}{\parbox{1.1cm}{\textbf{Multi-Frame}}} &
\multicolumn{3}{c}{\textbf{Kubric}} & \multicolumn{3}{c}{\textbf{DAVIS}} & \multicolumn{3}{c}{\textbf{Kinetics}} & \multicolumn{3}{c}{\textbf{RGB-Stacking}} \\ 
\cmidrule(lr){4-6} \cmidrule(lr){7-9} \cmidrule(lr){10-12} \cmidrule(lr){13-15} 
& & & AJ~$\uparrow$ & $<\delta^x_\textrm{avg}$~$\uparrow$ & OA~$\uparrow$ & AJ~$\uparrow$ & $<\delta^x_\textrm{avg}$~$\uparrow$ & OA~$\uparrow$ & AJ~$\uparrow$ & $<\delta^x_\textrm{avg}$~$\uparrow$ & OA~$\uparrow$ & AJ~$\uparrow$ & $<\delta^x_\textrm{avg}$~$\uparrow$ & OA~$\uparrow$ \\ 
\midrule
\parbox[t]{2mm}{\multirow{8}{*}{\rotatebox[origin=c]{90}{Supervised}}} 
& RAFT-C~\cite{teed2020raft} &  & $41.2$ & $58.2$ & $86.4$ & $30.7$ & $46.6$ & $80.2$ & $31.7$ & $51.7$ & $84.3$ & $42.0$ & $56.4$ & $91.5$ \\
& Kubric-VFS-Like~\cite{greff2022kubric} & & $51.9$ & $69.8$ & $84.6$ & $33.1$ & $48.5$ & $79.4$ & $40.5$ & $59.0$ & $80.0$ & $57.9$ & $72.6$ & $91.9$ \\
& RAFT-D~\cite{teed2020raft} & & $61.8$ & $79.1$ & $87.9$ & $34.1$ & $48.9$ & $76.1$ & $72.1$ & $85.1$ & $92.1$ & $50.6$ & $66.9$ & $85.5$ \\
& COTR~\cite{jiang2021cotr} & & $40.1$ & $60.7$ & $78.6$ & $35.4$ & $51.3$ & $80.2$ & $19.0$ & $38.8$ & $57.4$ & $6.8$ & $13.5$ & $79.1$ \\
& TAP-Net~\cite{doersch2022tap} & & $65.4$ & $77.7$ & $93.0$ & $38.4$ & $53.1$ & $82.3$ & $46.6$ & $60.9$ & $85.0$ & $59.9$ & $72.8$ & $90.4$ \\
& PIPs~\cite{harley2022particle} & \cmark & $59.1$ & $74.8$ & $88.6$ & $42.0$ & $59.4$ & $82.1$ & $35.3$ & $54.8$ & $77.4$ & $37.3$ & $51.0$ & $91.6$ \\
& TAPIR~\cite{doersch2023tapir} & \cmark & $84.7$ & $92.1$ & $95.8$ & $61.3$ & $73.6$ & $88.8$ & $57.2$ & $70.1$ & $87.8$ & $62.7$ & $74.6$ & $91.6$ \\
& CoTracker~\cite{karaev2023cotracker} & \cmark & $-$ & $-$ & $-$ & $64.8$ & $79.1$ & $88.7$ & $-$ & $-$ & $-$ & $-$ & $-$ & $-$ \\
 
\midrule
\parbox[t]{2mm}{\multirow{9}{*}{\rotatebox[origin=c]{90}{Self-supervised}}} 
& CRW-C~\cite{jabri2020space} & & $31.4$ & $48.1$ & $76.3$ & $7.7$ & $13.5$ & $72.9$ & $20.2$ & $33.6$ & $70.6$ & $25.3$ & $35.2$ & $70.1$ \\
& CRW-D~\cite{jabri2020space} & & $35.8$ & $52.4$ & $80.9$ & $23.6$ & $38.0$ & $77.2$ & $21.9$ & $36.8$ & $70.4$ & $13.1$ & $23.0$ & $83.4$ \\
& DIFT-C~\cite{tang2023emergent} & & $28.3$ & $45.2$ & $69.0$ & $18.1$ & $33.0$ & $68.8$ & $19.8$ & $33.7$ & $68.7$ & $13.2$ & $21.9$ & $56.3$ \\
& DIFT-D~\cite{tang2023emergent} & & $41.6$ & $59.8$ & $\bf{83.9}$ & $29.7$ & $48.2$ & $77.2$ & $19.5$ & $34.4$ & $70.1$ & $24.4$ & $38.9$ & $89.9$ \\
& Flow-Walk-C~\cite{bian2022learning} & & $49.4$ & $66.7$ & $82.7$ & $35.2$ & $51.4$ & $\bf{80.6}$ & $40.9$ & $55.5$ & $\bf{84.5}$ & $41.3$ & $55.7$ & $\bf{92.2}$ \\
& Flow-Walk-D~\cite{bian2022learning} & & $51.1$ & $68.1$ & $80.3$ & $24.4$ & $40.9$ & $76.5$ & $\bf{46.9}$ & $\bf{65.9}$ & $81.8$ & $\bf{66.3}$ & $\bf{82.7}$ & $91.2$ \\
& ARFlow-C~\cite{liu2020learning} & & $52.3$ & $68.1$ & $81.4$ & $35.0$ & $51.8$ & $79.7$ & $27.3$ & $44.3$ & $79.5$ & $33.0$ & $47.2$ & $91.9$ \\
\cdashline{2-15}
& Ours - \ourmodel-C & & $\bf{54.2}$ & $\bf{72.4}$ & $82.6$ & $\bf{41.8}$ & $\bf{60.9}$ & $78.3$ & $31.9$ & $52.3$ & $72.9$ & $39.8$ & $56.5$ & $90.8$ \\
& Ours - \ourmodel-D & & $51.4$ & $71.7$ & $\bf{83.9}$ & $30.3$ & $49.4$ & $77.3$ & $36.3$ & $59.2$ & $71.0$ & $56.4$ & $74.1 $ & $90.9$ \\
\bottomrule
& *FlowWalk-C (hi-res)~\cite{bian2022learning} &  & $66.1$ & $82.6$ & $88.6$ & $45.9$ & $63.4$ & $80.8$ & $41.5$ & $60.7$ & $80.9$ & $34.3$ & $48.8$ & $89.8$ \\
& *RAFT-C (hi-res)~\cite{teed2020raft} &  & $66.7$ & $82.8$ & $88.5$ & $42.6$ & $61.0$ & $80.4$ & $39.4$ & $58.2$ & $80.6$ &  $47.2$ & $62.4$ & $91.6$  \\
\bottomrule
\end{tabular}
}}%
\vspace{0.1in}
\caption{Comparison of our method and baselines on the Tap-Vid benchmark. We show strong performance on all four Tap-Vid benchmarks. We outperform self-supervised methods on Kubric and DAVIS and are comparable with several supervised methods. * FlowWalk-C and RAFT-C performs better than several supervised tracking baselines when evaluted at higher-resolution (close to their training resolution).}
\label{tab: main_results}
\end{table*}

\begin{table}[tbp]
\centering 
{\small
\setlength{\tabcolsep}{4pt}
\begin{tabularx}{0.9\linewidth}{X ccc ccc} 
\toprule
\multirow{2}{*}{\textbf{Method}} & \multicolumn{3}{c}{\textbf{Kubric}} & \multicolumn{3}{c}{\textbf{DAVIS}} \\
\cmidrule(lr){2-4} \cmidrule(lr){5-7} 
& AJ~$\uparrow$ & $<\delta^x_\textrm{avg}$~$\uparrow$ & OA~$\uparrow$ & AJ~$\uparrow$ & $<\delta^x_\textrm{avg}$~$\uparrow$ & OA~$\uparrow$ \\
\midrule
Supervised & $63.7$ & $83.2$ & $83.9$ & $39.1$ & $59.6$ & $77.3$ \\
\midrule
CRW & 25.4 & 39.3 & 83.3 & 10.4 & 19.2 & 76.9 \\ 
+ Label Warping & 45.3 & 62.2 & 83.1 & 32.1 & 48.9 & 78.2 \\  
+ Smoothness loss & 49.0 & 66.7 & 84.4 & 33.0 & 50.5 & 79.4 \\ 
+ Train stride $s=2$ & 47.7 & 65.6 & 84.4 & 34.5 & 52.1 & 79.3 \\  
Trained w/ Kinetics  & 47.5 & 65.0 & 83.8 & 34.6 & 52.6 & 78.7 \\  
\midrule
Eval stride $s=4$ & 37.8 & 53.8 & 84.1 & 23.5 & 38.4 & 78.8 \\
Eval stride $s=2$ & 47.5 & 65.0 & 83.8 & 34.6 & 52.6 & 78.7 \\  
Eval stride $s=1$ & 54.2 & 72.4 & 82.6 & 41.8 & 60.9 & 78.3 \\

\bottomrule
\end{tabularx}}
\vspace{2mm}
\caption{{\bf Model variations}. We evaluate several variations of our model. We consider a version of the architecture trained via supervised learning, ablated versions of the model, a version trained on ``in-the-wild'' Kinetics~\cite{kay2017kinetics} videos, and several different stride configurations during evaluation. } %
\label{tab:model_ablations}
\end{table}

\subsection{Comparison to other methods}
In Table~\ref{tab: main_results}, we compare our method to previously proposed self-supervised tracking and supervised approaches. %

\xhdr{Baselines.} We compare our method with several self-supervised baselines. Following Wang et al.~\cite{wang2023tracking}, we run all self-supervised methods in Chained (-\textbf{C}) and Direct (-\textbf{D}) settings and show their performance in Table~\ref{tab: main_results} (bottom section). In the chained setting, the predictions are made for adjacent frames and they are chained together over time to form long-range tracks. In the direct setting, the query point frame is compared directly with all frames and the motion is computed for each pair independent of time. For all self-supervised methods (none of which explicitly predict occlusion masks), we test for occlusion using cycle consistency over the predicted tracks using the same approach used in our method. We evaluate the following methods: %
\vspace{2mm}
\begin{itemize}[label=$\bullet$,itemsep=0pt,topsep=0pt,leftmargin=5mm]
    \item{\textbf{Diffusion Features (DIFT)}: Next, we consider recent work that uses diffusion features~\cite{tang2023emergent} for tracking as a baseline. This model extracts features from pretrained Stable Diffusion model~\cite{rombach2022high} and Ablated Diffusion~\cite{dhariwal2021diffusion} model, and successfully uses them for a variety of correspondence tasks. For real images, they add a noise of a specific time step $t$, feed it to the network together with $t$ to extract intermediate layer activations. These activations are then used as features. We use the Ablated Diffusion variation of the model, since it performs better for temporal correspondence tasks. }
    \item{\textbf{Contrastive Random Walk (CRW)}: We first evaluate the  CRW model of Jabri et al.~\cite{jabri2020space} on the TapVid benchmarks. In CRW, the images are partitioned into small patches, and the model is trained to learn representations for these patches through cycle consistency. The model is trained for cycle consistent videos of length 10 and sub-cycles are also supervised. Since the CRW operates at the patch level, the correspondences computed are very coarse.}
    \item{\textbf{FlowWalk}: We also evaluate FlowWalk~\cite{bian2022learning} on these tracking benchmarks, using the reported results from OmniMotion~\cite{wang2023tracking}. FlowWalk is another contrastive random walk-based method that is trained for cycle consistency at multiple scales, and which can perform optical flow estimation. The output from their model is pixel-level flow values.} %
    \item{\textbf{ARFlow}: We train ARFlow~\cite{liu2020learning} on the TapVid-Kubric dataset and evaluate it on the TapVid benchmarks. ARFlow is an unsupervised optical flow method which uses spatial transformations to provide self-supervision in training.} %
    \item{\textbf{Supervised methods:} We also show several supervised methods on the TapVid benchmark in the top section of Table~\ref{tab: main_results}. Out of these, PIPs, TAPIR, and CoTracker are trained with multiple frames and use local, spatial-temporal information to refine their tracks. }
\end{itemize}

\xhdr{Hi-res flow evaluation.} TapVid~\cite{doersch2022tap} and OmniMotion~\cite{wang2023tracking} evaluate optical flow methods at 256 $\times$ 256 resolution. We find that evaluating flow methods at lower resolutions when they were originally trained at higher-resolutions (e.g. 384 $\times$ 512, 448 $\times$ 1024), results in low performance. We evaluate them at 512 $\times$ 512 (closer to their training resolution) by upsampling the input images and find that they are competitive with several tracking methods designed for the TapVid benchmark.

\xhdr{Quantitative comparisons.} We compare our method qualitatively to self-supervised baselines in Table~\ref{tab: main_results}. We find that it outperformed other self-supervised baselines on TapVid-Kubric and TapVid-DAVIS and have competitive numbers with several supervised methods, such as TAP-Net. We evaluate our method at a stride $s=1$ for chained setting and $s=2$ for direct setting, except for the TapVid-RGB-Stacking benchmark, where we find the best performance at $s=4$ (perhaps because the highly synthetic nature of the videos leads to local ambiguity). Among other self-supervised methods, CRW operates at the patch level, so the correspondence we get is very coarse and does not work well when evaluated for tracking. DIFT outperforms CRW, but fails to outperform other self-supervised methods, possibly because DIFT's text-to-image diffusion features rely on the semantics of the image to find correspondence, but tracking requires relying on low-level motion cues to find correspondences.

\begin{figure*}[tp]
    \centering
    \includegraphics[width=\textwidth]{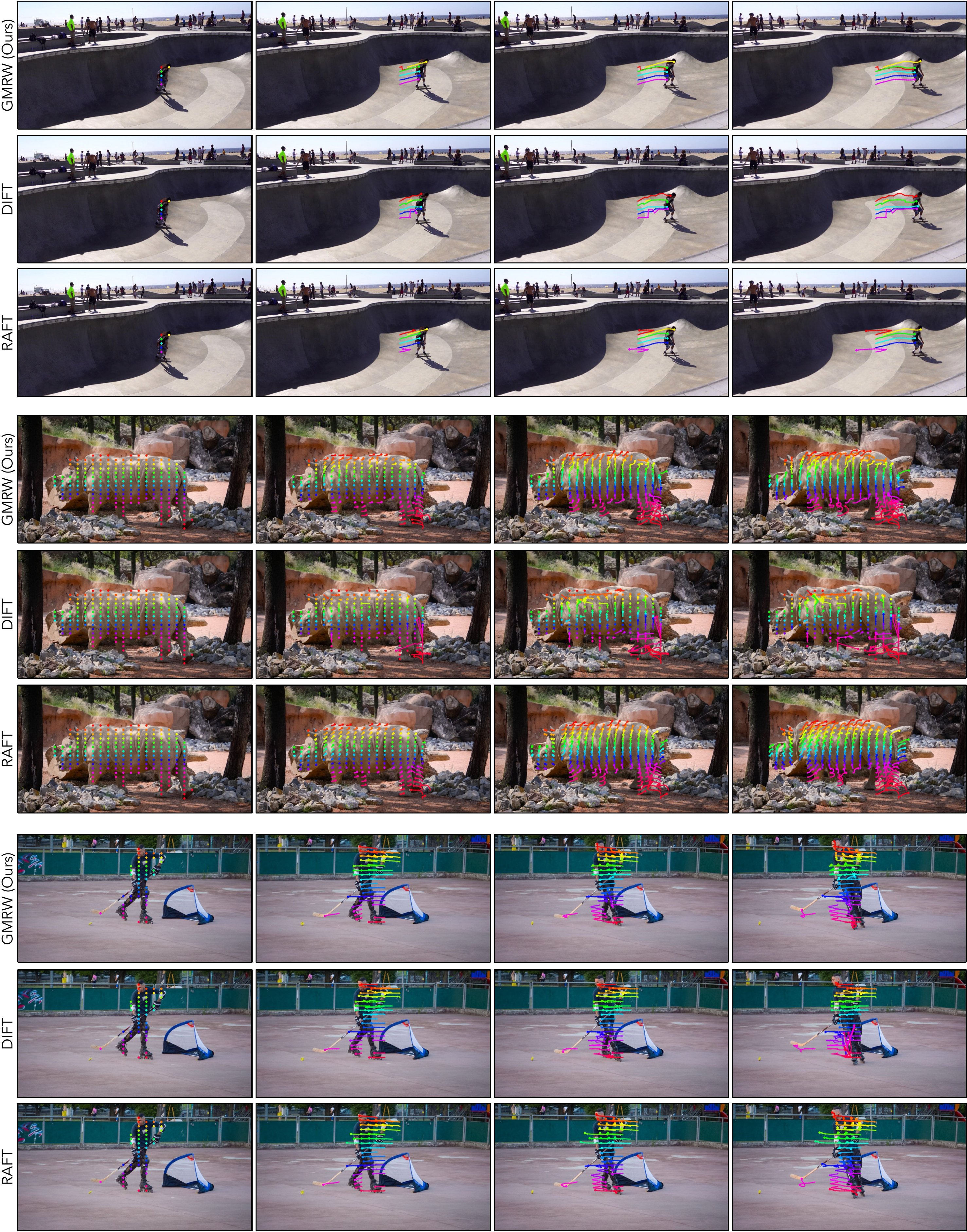}
    \caption{{\bf Qualitative results.} We show qualitative results for TapVid-DAVIS videos and compare them with DIFT and RAFT. DIFT relies on semantic correspondences and often loses the point of interest when motion occurs in the video. RAFT produces accurate movements for several tracks but suffers from drifting of points when the predictions are chained over a long period. In the first video, our method can track points accurately over the long timesteps. RAFT, on the other hand, loses current locations for 2 query points and latches on points on the ground and starts tracking them. In the other 2 videos as well, our method works better than RAFT and DIFT. DIFT produces inaccurate tracks that do not capture motion well. RAFT being accurate most of the time, loses track of points close to the boundary. }
    \label{fig:qual_results}
\end{figure*}

\begin{figure*}[tp]
    \centering
    \includegraphics[width=0.86\textwidth]{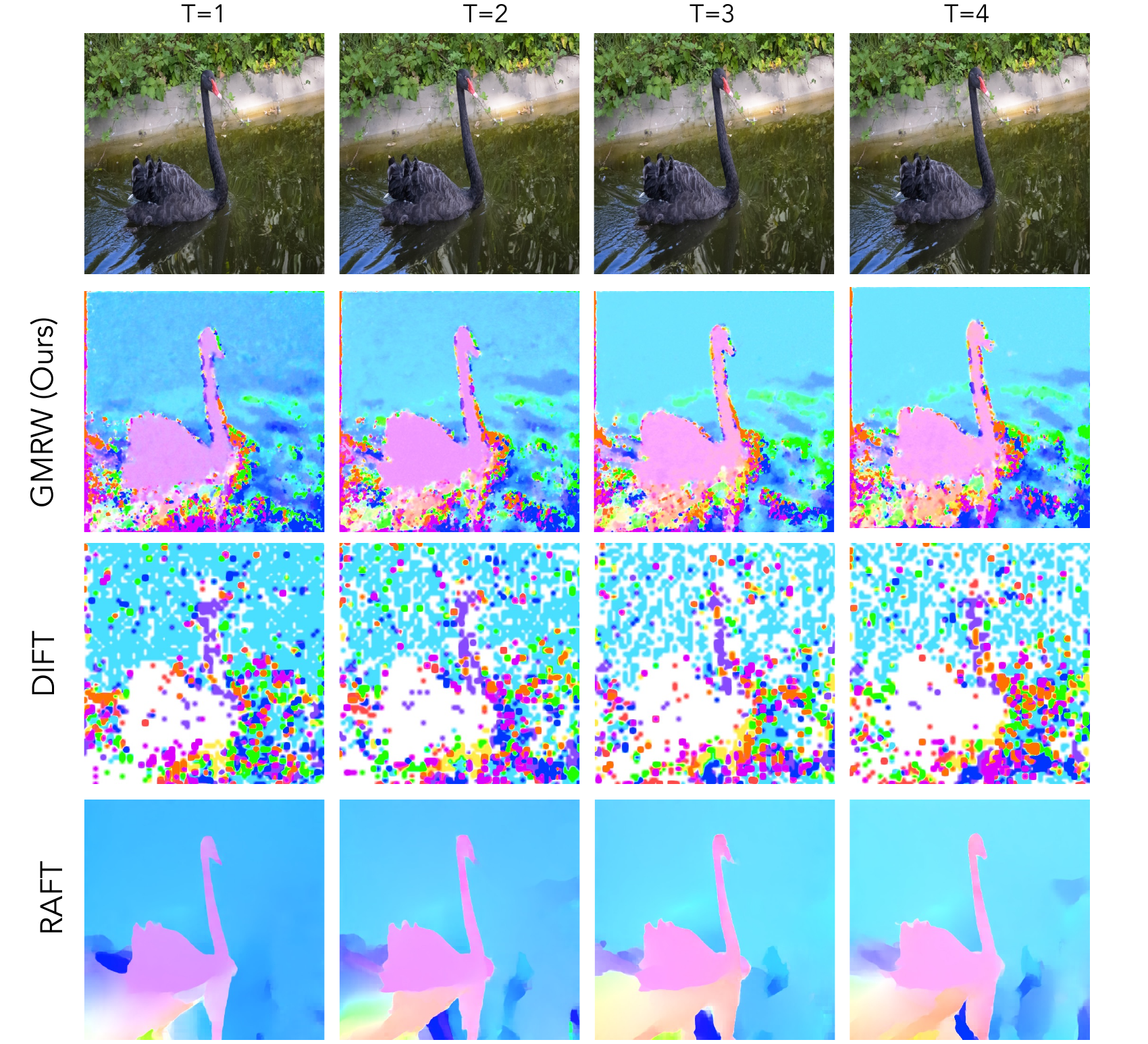}
    \caption{{\bf Optical flow visualization.} Although our method is not trained for the optical flow prediction task, it is able to produce reasonable flow outputs over multiple timesteps. RAFT produces high quality flows as it is an optical flow method trained for this objective. DIFT predicts inaccurate flow which are spotty in nature, suggesting that it relies on finding semantic correspondence for certain points in the image, instead of relying local motion cues.}
    \label{fig:qual_results_flow}
\end{figure*}

\xhdr{Qualitative results.} We show qualitative results on tracking in Figure~\ref{fig:qual_results} and optical flow in Figure~\ref{fig:qual_results_flow} and compare them with DIFT and RAFT. Our method is able to produce reasonable tracks for long time horizon and works better than DIFT and RAFT. Even though our method is not supervised for the optical flow task, it is still able to produce reasonable-quality flow maps. 

\subsection{Model variations and ablations}
We investigate our model's components in Table~\ref{tab:model_ablations}. First, we distinguish between the performance of our architecture versus our learning procedure. To test this, we train our global matching-based architecture (Sec.~\ref{sec:architecture}) by training a {\em supervised} variant of our model (evaluated at stride $s=4$). Our training setup closely follows TAP-Net~\cite{doersch2022tap}. We randomly sample 2 frames from a TapVid-Kubric training video and supervise the estimated motion (Eq.~\ref{eq:motion}) using the ground truth, using a scaled Huber loss. We compare with TAP-Net as it is modeled with a similar setup as ours without multi-scale features and spatial-temporal iterative refinement.  We see that the performance of our supervised model is better than TAP-Net on TAPVid-Kubric and TAPVid-DAVIS. These results suggest that the architecture is capable of obtaining tracking results that are on par with other supervised architectures that have been proposed for TAP.

Next, we evaluate the different components of our approach. We train our self-supervised model on the Kubric dataset without label warping, finding that it performs poorly. This suggests that it can find shortcuts using positional embeddings without learning meaningful representations. After adding label warping, we see a large boost in performance. Next, we train our model with smoothness loss in addition to cycle consistency and see a small improvement in performance. Interestingly, this is in contrast to Bian et al.~\cite{bian2022learning}, which found the smoothness loss to be critically important. We hypothesize that this is due to our use of global matching, rather than the coarse-to-fine search used in~\cite{bian2022learning}. The latter may implicitly require neighboring pixels to match to similar locations, since the finer scales are obtained by warping a feature map~\cite{brox2004high} using the estimated optical flow of each pixel's neighbors. This highlights a potential advantage of our network architecture. 

Next, to test the effect of stride in training, we lower the training stride from $s=4$ to $s=2$ and see that we obtain a minor performance improvement. Finally, we trained our model with all components on the Kinetics 400 dataset and see that performance improves slightly on TAPVid-DAVIS, while decreasing for the TAPVid-Kubric dataset. This suggests, first, that our model can successfully be trained using unlabeled in-the-wild video, rather than the synthetic datasets used in existing supervised learning work. Second, since the improvement was for a dataset of real videos (DAVIS) and decreased performance was for a synthetic dataset (Kubric), this may suggest that it is beneficial to match the distribution by training on real video. %

\section{Conclusion}
In this paper, we present a simple, self-supervised model for addressing long-range tracking for the Tracking Any Point task.  We adapt the global matching transformer architecture to training via the contrastive random walk. Our approach significantly outperforms previous self-supervised approaches on the TAP-Vid benchmark, obtaining performance that on some metrics is close to that of supervised TAP-Net~\cite{doersch2022tap}. While our model can handle occlusions to some degree using cycle-consistency, we did not design our model to handle occlusion explicitly. It also does not have the capability to track pixels through occlusions. We see this work as a step towards building tracking methods trained using self-supervised learning, and to creating computer vision models that learn at scale from unlabeled data.

\xhdr{Acknowledgements.} This work was supported by Toyota Research Institute and Cisco Systems. We would like to thank Adam Harley, Allan Jabri, Daniel Geng and Ziyang Chen for helpful discussions and feedback on the paper.

\bibliographystyle{eccv2024template/splncs04}
\bibliography{main}
\end{document}